\documentclass[conference]{IEEEtran}
\IEEEoverridecommandlockouts
\usepackage{cite}
\usepackage{amsmath,amssymb,amsfonts}
\usepackage{algorithmic}
\usepackage{graphicx}
\usepackage{textcomp}
\usepackage{xcolor}
\usepackage{multirow}
\usepackage{mathrsfs}
\def\BibTeX{{\rm B\kern-.05em{\sc i\kern-.025em b}\kern-.08em
    T\kern-.1667em\lower.7ex\hbox{E}\kern-.125emX}}

\begin{document}
\title{SOC: hunting the underground inside story of the ethereum Social-network Opinion and Comment}
\author{TonTon Hsien-De~Huang$^\dagger$$^*$,
        Po-Wei~Hong$^\dagger$$^\delta$,        
        Ying-Tse~Lee$^\dagger$,
        Yi-Lun~Wang$^\dagger$,
        Chi-Leong~Lok$^\dagger$,        
		and~Hung-Yu Kao$^*$\\
$^\dagger$Cyber-Security and AI Research Lab., Leopard Mobile Inc. (Cheetah Mobile Taiwan Agency), Taiwan\\
$^\delta$Department of Computer Science and Information Engineering, National Cheng Kung University, Taiwan\\
$^*$Department of Computer Science and Engineering, National Tsing Hua University, Taiwan\\
TonTon@TWMAN.ORG\\
}


\markboth{Journal of \LaTeX\ Class Files,~Vol.~14, No.~8, August~2015}%
{Shell \MakeLowercase{\textit{et al.}}: Bare Demo of IEEEtran.cls for IEEE Communications Society Journals}

\maketitle

\begin{abstract}
The cryptocurrency is attracting more and more attention because of the blockchain technology. Ethereum is gaining a significant popularity in blockchain community, mainly due to the fact that it is designed in a way that enables developers to write smart contracts and decentralized applications (Dapps). There are many kinds of cryptocurrency information on social network. The risks and fraud problems behind it have pushed many countries including the United States, South Korea, and China to make warnings and set up corresponding regulations. However, the security of Ethereum smart contracts has not gained much attention. Through the Deep Learning approach, we propose a method of sentiment analysis for Ethereum's community comments.

In this research, we first collected the users’ cryptocurrency comments from the social network and then fed to our LSTM + CNN model for training. Then we made prediction through sentiment analysis. With our research result, we have demonstrated that both the precision and the recall of sentiment analysis can achieve 0.80+. More importantly, we deploy our sentiment analysis\footnote{https://www.ratingtoken.net/sentiment/} on RatingToken and Coin Master (mobile application of Cheetah Mobile Blockchain Security Center\footnote{https://play.google.com/store/apps/details?id=coinmaster.blockchain.assets.\\holding}\footnote{https://itunes.apple.com/app/apple-store/id1377543556}). We can effectively provide the detail information to resolve the risks of being fake and fraud problems.


\end{abstract}

\section{Introduction}
Since Satoshi Nakamoto published the article "Bitcoin: A Peer-to-Peer Electronic Cash System" in 2008 \cite{bitcoin}, and after the official launch of Bitcoin in 2009, technologies such as blockchain and cryptocurrency have attracted attention from academia and industry. At present, the technologies have been applied to many fields such as medical science, economics, Internet of Things \cite{surveyvulnerability}. Since the launch of Ethereum (Next Generation Encryption Platform) \cite{BitcoinMagazine} with smart contract function proposed by Vitalik Buterin in 2015, lots of attention has been obtained on its dedicated cryptocurrency Ether, smart contract, blockchain and its decentralized Ethereum Virtual Machine (EVM). The main reason is that its design method provides developers with the ability to develop Decentralized apps (Dapps), and thus obtain wider applications. A new application paradigm opens the door to many possibilities and opportunities.

Initial Coin Offerings (ICOs) is a financing method for the blockchain industry. As an example of financial innovation, ICO provides rapid access to capital for new ventures but suffers from drawbacks relating to non-regulation, considerable risk, and non-accountability. According to a report prepared by Satis Group Crypto Research, around 81\% of the total number of ICOs launched since 2017 have turned out to be scams \cite{Pennsylvania}. Also, according to the inquiry reveals of the University of Pennsylvania that many ICO failed even to promise that they would protect investors against insider self-dealing\footnote{https://www.ccn.com/81-of-icos-are-scams-u-s-losing-token-sale-market-share-report/}. 

Google\footnote{https://support.google.com/adwordspolicy/answer/7648803}, Facebook\footnote{https://www.facebook.com/business/news/new-ads-policy-improving-integrity-and-security-of-financial-product-and-services-ads}, and Twitter\footnote{https://www.cnbc.com/2018/03/26/twitter-bans-cryptocurrency-advertising-joining-other-tech-giants-in-crackdown.html} have announced that they will ban advertising of cryptocurrencies, ICO etc. in the future. Fraudulent pyramid selling of virtual currency happens frequently in China. The People’s Bank of China has banned the provision of services for virtual currency transactions and ICO activities\footnote{https://www.ethnews.com/btcchina-exchange-to-halt-trading-in-china}. 

The incredibly huge amount of ICO projects make it difficult for people to recognize its risks. In order to find items of interest, people usually query the social network using the opinion of those items, and then view or further purchase them. In reality, the opinion of social network platforms is the major entrance for users. People are inclined to view or buy the items that have been purchased by many other people, and/or have high review scores.

Sentiment analysis is contextual mining of text which identifies and extracts subjective information in the source material and helping a business to understand the social sentiment of their brand, product or service while monitoring online conversations. To resolve the risks and fraud problems, it is important not only to analyze the social-network opinion, but also to scan the smart contract vulnerability detection.

\section{Our Proposed Methodology}
\begin{figure}[htbp]
	\centering
	\includegraphics[width=3.6in,height=3.1in]{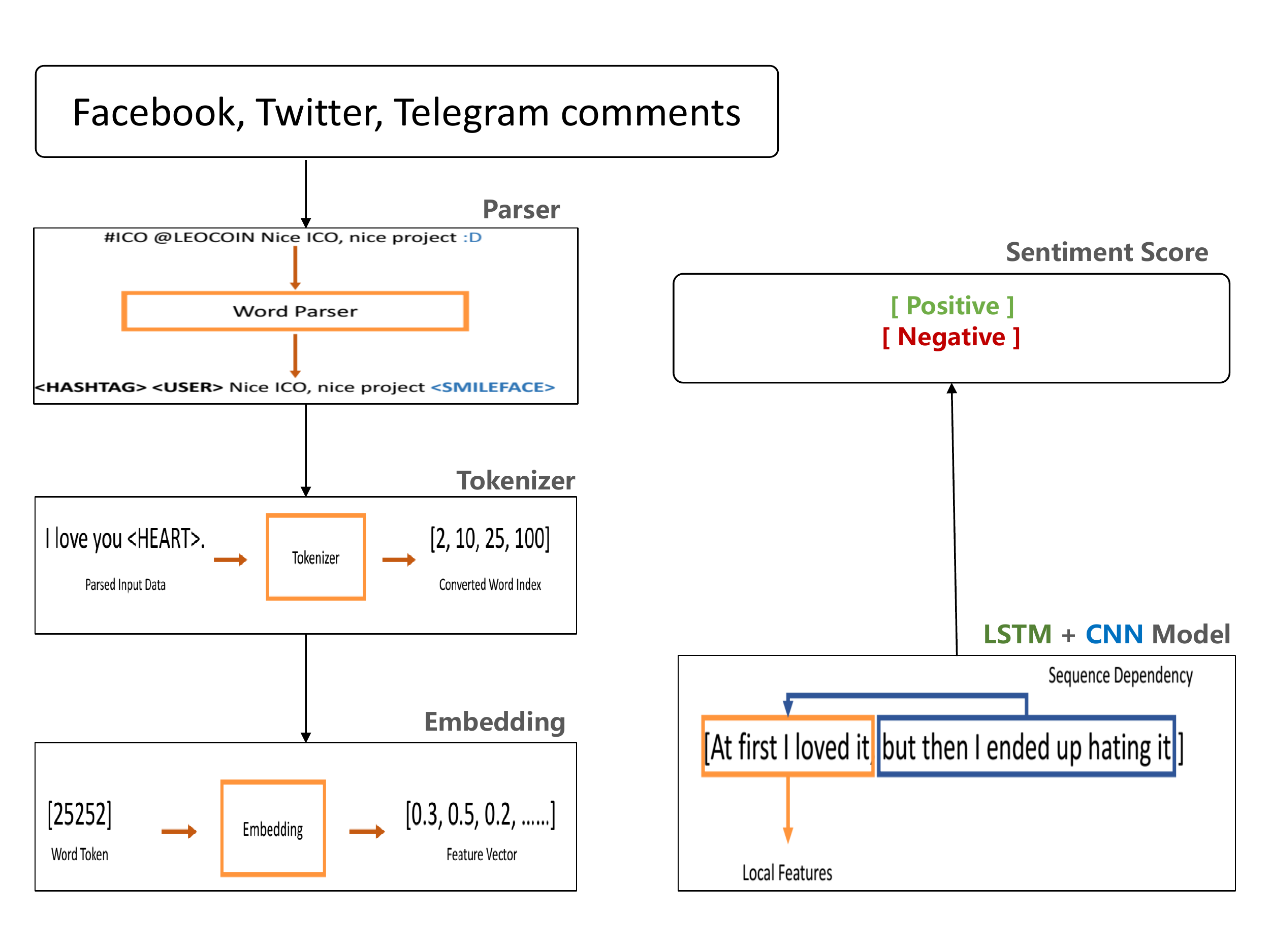}
	\caption{Our system flow-chart.}\label{fig: SOC001}
\end{figure}

\begin{figure}[htbp]
	\centering
	\includegraphics[width=3.6in,height=3.1in]{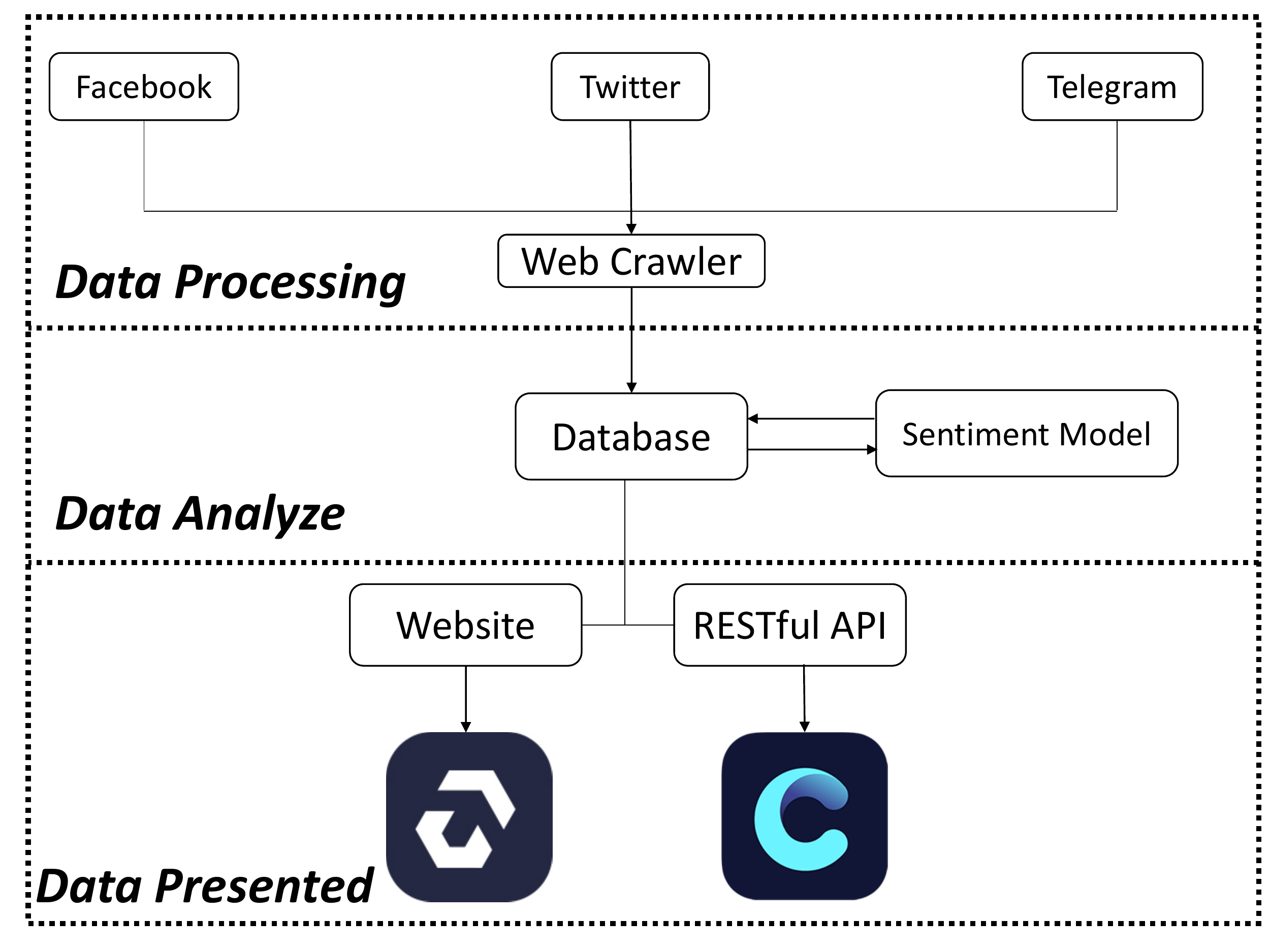}
	\caption{Our system architecture.}\label{fig: SOC006}
\end{figure}

We proposed two methodologies which integrate Long Short Term Memory network (LSTM) and Convolutional Neural Network (CNN) into sentiment analysis model, one outputs the softmax probability over two kinds of emotions, positive and negative. The other model outputs the tanh sentiment score of input text ranged [-1, 1], -1 represents the negative emotion and vice versa. Fig. \ref{fig: SOC001} shows our system flow-chart and Fig. \ref{fig: SOC006} is our system architecture. Detail descriptions are explained below.

\textbf{Tokenizing and Word Embedding.} The raw input text can be noisy. They contain specific words which could affect the model training process. To clean these input text, we use the tokenizer from Stanford NLP \cite{glove} to remove some unnecessary tokens such as username, hashtag and URL. Word embedding is a distributed representation of a word, which is suitable for the input of neural networks. In this work, we choose the word embedding size d = 100, and use the 100-dimensional GloVe word embeddings pre-trained on the 27B Twitter data to initialize the word embeddings. 

\begin{figure}[htbp]
	\centering
	\includegraphics[width=3.6in,height=3in]{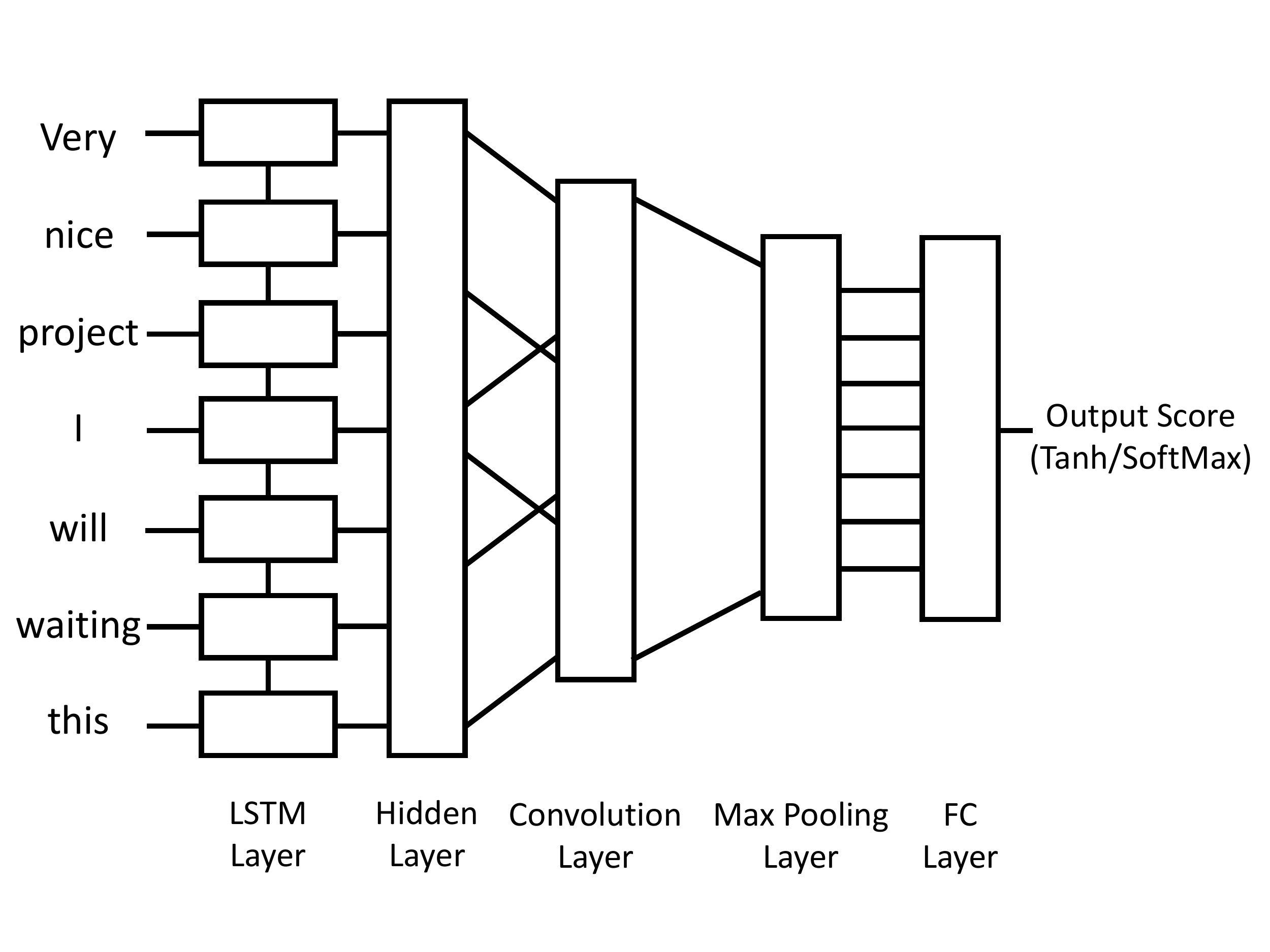}
	\caption{Our LSTM + CNN architecture.}\label{fig: model}
\end{figure}

\textbf{Model Architecture.} The architecture of our models are shown in Figure \ref{fig: model}, both based on the combination of LSTM and CNN.
\begin{itemize}
\item Long Short-Term Memory network (LSTM): LSTM \cite{LSTM} is a type of RNNs to solve the gradient vanishing problems of RNNs. Recurrent neural networks (RNNs) have been effective in handling time-series data. The calculation history is stored in recurrent hidden units which dependent on the previous hidden unit. In this work, LSTM is used to extract sentiment features in a contextual essay. 
\item Convolutional Neural Network (CNN): CNN is composed of hidden layers, fully connected layers, convolution layers, and pooling layers. The hidden layers are used to increase the complexity of the model. If the same number of neural is associated with the input image, the number of parameters can be significantly reduced, adapting to the function structure much properly. In this work, CNN is used to capture local sentiment features.
\item Activation Functions: Scaled Exponential Linear Unit (SELU) \cite{SELU} is a variation of the Exponential Linear Unit (ELU) for creating self-normalizing neural networks, which can be represented as two hidden layers using selu as activation function in this work.
\end{itemize}

\begin{figure}[hbtp]
	\centering
	\includegraphics[width=3.5in,height=3in]{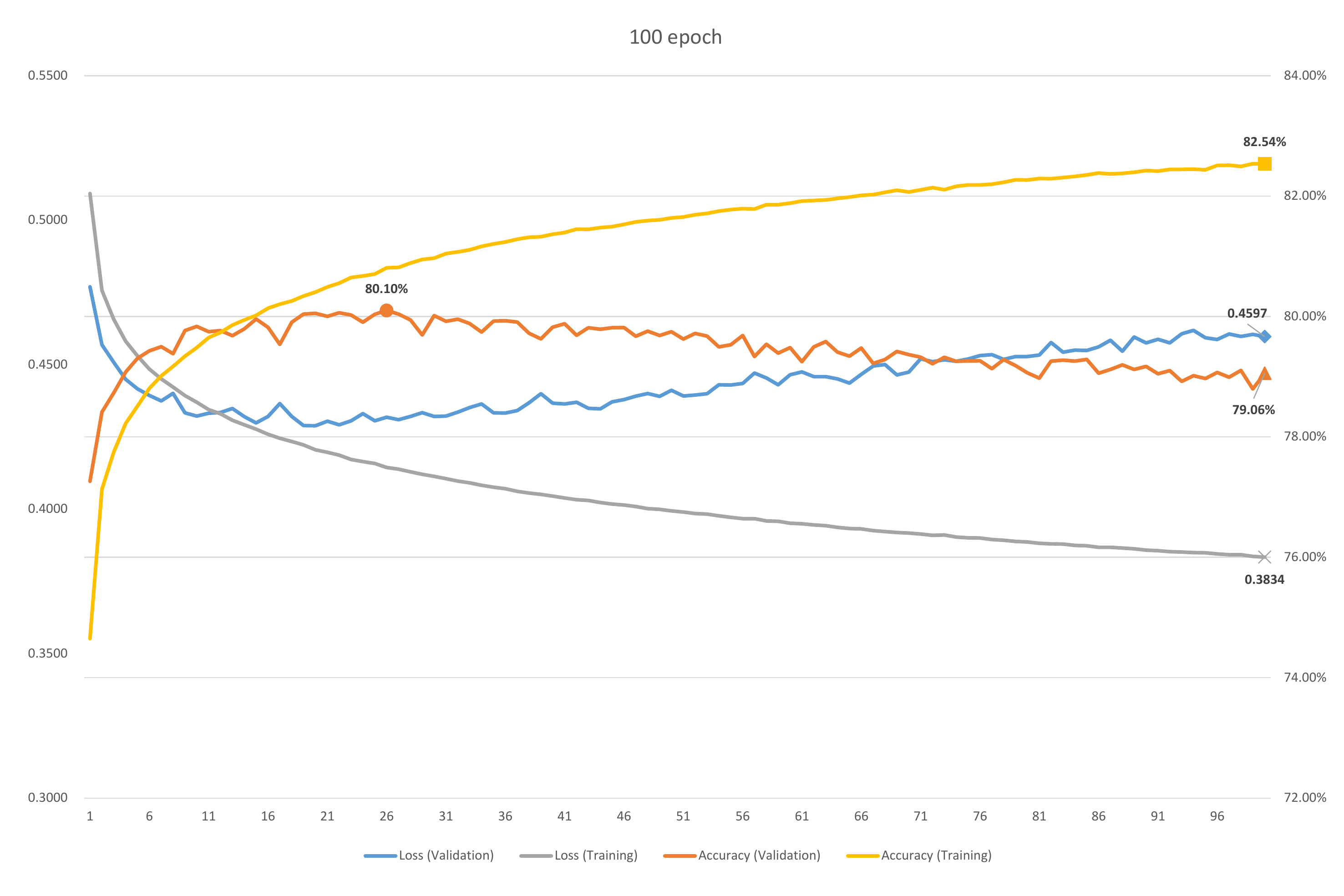}
	\caption{100 epoch result.}\label{fig: SOC005}
\end{figure}

\textbf{Training.} We use a max input sequence length of 64 during training and testing. For LSTM layer, the number of hidden units is 64, and all hidden layer size is 128. All trainable parameters have randomly initialized. Our model can be trained by minimizing the objective functions, we use $\ell_{2}$ function for tanh model and cross-entropy for softmax model. For optimization, we use Adam \cite{Adam} with the two momentum parameters set to 0.9 and 0.999 respectively. The initial learning rate was set to 1E-3, and the batch size was 2048. Fig. \ref{fig: SOC005} is our experiment result.

\textbf{Ranking.} The design of the model can guarantee the accuracy of the sentiment score of each text. However, when calculating the project score, problems occur when projects have the same score. For example, when the score of item A and item B are the same, while item A has 1000 texts, and item B has only 1 text. If it is only a simple calculation of the weighted score, it does not reflect the difference in the difference between the amount of data between Project A and B. So we designed a score-based scoring algorithm (as shown in the equation \ref{ranking}).

\begin{equation}
\label{ranking}
	\begin{split}
		& \textup{Let } N \textup{ be the set of all tokens,} \\
		& C_n \textup{ be the number of comments of day n,} \\
		& \textup{Given a certain token } X\in N, \\
		& \textup{First calculate the total number of comment } M_x \\
		& \textup{of a time window } T_i, T_{i+1}, ..., T_j \\
		& M_x = \sum_{t=i}^{j}C_t \\
		& \textup{And calculate the score weight } W_x \textup{ of given token.} \\
		& W_x = \frac{M_x}{max_{k:k\in N}(M_k)} \\
		& \textup{Then we can calculate the adjusted score by applying } \\
		& \textup {score weight.} \\
		& Score_{x}^{adj}= Score_{x}^{orig}\ast W_{x} \\
	\end{split}
\end{equation}

\section{Datasets and Experimental Results}
We train our models on Sentiment140 and Amazon product reviews. Both of these datasets concentrates on sentiment represented by a short text. Summary description of other datasets for validation are also as below:

\begin{itemize}
\item Sentiment140\footnote{https://www.kaggle.com/kazanova/sentiment140/kernels}: This is our trainging dataset which contains 1.6 million tweets extracted using the twitter api. The tweets have been annotated (0 = negativity, 2 = objectivity, 4 = positivity)\cite{sentiment140}.
\item Amazon product reviews\footnote{http://jmcauley.ucsd.edu/data/amazon/}: This dataset contains product reviews and metadata from Amazon, including 142.8 million reviews spanning May 1996 - July 2014. This dataset includes reviews (ratings, text, helpfulness votes), product metadata (descriptions, category information, price, brand, and image features), and links (also viewed/also bought graphs).
\item SST-1\footnote{https://www.cs.cornell.edu/people/pabo/movie-review-data/}: Stanford Sentiment Treebank: an extension of Movie Reviews but with train/dev/test splits provided and fine-grained labels (very positive, positive, neutral, negative, very negative), re-labeled by Socher et al.
\item SST-2: Same as SST-1 but with neutral reviews removed and binary labels \cite{kim}.
\item SWN-1\footnote{https://sentiwordnet.isti.cnr.it/}: SentiWordNet is a lexical resource for opinion mining. SentiWordNet assigns to each synset of WordNet three sentiment scores: positivity, negativity, objectivity\cite{SWN}.
\item SWN-2: Remove objective data. Score calculate method is "positive and negative".
\item ACL\footnote{https://www.kaggle.com/rahulin05\/sentiment-labelled-sentences-data-set\#imdb\_labelled.txt}: The Data has sentences from 3 sources IMDB reviews, Yelp reviews and Amazon cell phones and accessories reviews. Each line in Data Set is tagged positive or negative.
\end{itemize}

We have provided baseline results for the accuracy of other models against datasets (as shown in Table \ref{tab: tab1}) For training the softmax model, we divide the text sentiment to two kinds of emotion, positive and negative. And for training the tanh model, we convert the positive and negative emotion to [-1.0, 1.0] continuous sentiment score, while 1.0 means positive and vice versa. We also test our model on various models and calculate metrics such as accuracy, precision and recall and show the results are in Table \ref{tab: tab2}. Table \ref{tab: tab3}, Table \ref{tab: tab4}, Table \ref{tab: tab5}, Table \ref{tab: tab6} and Table \ref{tab: tab7}. Table \ref{tab: tab8} are more detail information with precisions and recall of our models against other datasets.

From the above experimental results, it can be found that the dataset from IMDB and YELP do not have neutral problem, so it can be solved with softmax. However, when we compare SST with SWN, we found many neutral sentences inside, which causes Softmax achieved poor results. Then we tried to use tanh, it got better results, but because of the data set, the current effect is still lower than other papers, but it is already better than using softmax! Therefore, we also tried to introduce the concept of emotional continuity in this work, we express the emotion as the output space of tanh function [-1, 1], and -1 represents the most negative emotion, +1 Representing the most positive emotions, 0 means neutral sentences without any emotions. Compared to other dichotomies or triads, we think this is more intuitive. We have listed the scores using the Yelp dataset as an example and divided them into positive (0.33, 1), neutral [-0.33, 0.33] and negative [-1, 0.33) emotions and each three representative sentences were presented in Table \ref{tab: tab9} (the score is rounded to 5 decimal places).

In order to validate  the concept of emotional continuity, we re-deploy the rating of the product of cell phones and accessories reviews (Amazon product reviews) from five sentiment scores (5: very positive, 4: positive, 3: neutral, 2: negative, 1: very negative) to three sentiment scores ((1, 0.33): very positive and positive, [0.33, -0.33]: neutral, {-0.33, -1]: negative and very negative) and re-train our model. We test our tanh and softmax model on other Amazon product reviews datasets (musical instruments, home and kitchen and toys and games etc... ) and calculate metrics such as accuracy of neutral reviews and all reviews. With this experimental results, we found tanh model got better results and can be solved neutral problem. We show the results are in Table \ref{tab: tab10}. 

Our system run on a 64-bit Ubuntu 14.04, and hardware setting are 128 GB DDR4 2400 RAM and Intel(R) Xeon(R) E5-2620 v4 CPU, NVIDIA TITAN V, TITAN XP, and GTX 1080 GPUs; the software setting is the nvidia-docker tensorflow:18.04-py3 on NVIDIA cloud. More specifically, our training models run on a single GPU (NVIDIA TITAN V), with roughly a runtime of 5 minutes per epoch. We run all the models up to 300 epochs and select the model which has the best accuracy on the testing set.

\begin{table}
\begin{center}
\caption{The accuracy of other models against datasets.}\label{tab: tab1}
\begin{tabular}{|c|c|c|c|c|c|}
\hline
Model & SST-1 & SST-2 & IMDB & Yelp2013 & Yelp2014 \\ \hline
CNN-rand & 45.0\% & 82.7\% & - & - & -\\ \hline
CNN-static & 45.5\% & 86.8\% & - & - & -\\ \hline
CNN-non-static & 45.0\% & 87.2\% & - & - & - \\ \hline
CNN-multichannel & 47.4\% & 88.1\% & - & - & -\\ \hline
NSC & - & - & 44.30\% & 62.70\&\% & 63.70\%\\ \hline
NSC + LA & - & - & 48.70\% & 63.10\% & 63.0\%\\ \hline
NSC + UPA & - & - & 53.30\% & 65.0\% & 66.70\% \\ \hline
\end{tabular}
\end{center}
\end{table}

\begin{table}
\begin{center}
\caption{The accuracy of our models against other datasets.}\label{tab: tab2}
\begin{tabular}{|c|c|c|c|c|}
\hline
\multirow{2}{*}{} \textbf{Our Models} & \multicolumn{4}{c|}{\textbf{Testing Datasets}}                                             \\ \cline{2-5} 
     \textbf{(Training Datasets)}  &        \textbf{\textit{S-140}}      &          \textbf{\textit{ACL}}      &         \textbf{\textit{IMDB}}     &      \textbf{\textit{YELP}}     \\ \hline
\multirow{2}{*}{} Tanh & \multirow{2}{*}{78.69\%} & \multirow{2}{*}{79.40\%} & \multirow{2}{*}{78.80\%} & \multirow{2}{*}{82.90\%} \\
                  (S-140) &                   &                   &                   &                   \\ \hline
\multirow{2}{*}{} Tanh & \multirow{2}{*}{81.85\%} & \multirow{2}{*}{78.50\%} & \multirow{2}{*}{73.30\%} & \multirow{2}{*}{82.40\%} \\
         (S-140 + Amazon) &                   &                   &                   &                   \\ \hline
\multirow{2}{*}{}  \textbf{Our Models} & \multicolumn{4}{c|}{\textbf{Testing Datasets}}                                              \\ \cline{2-5} 
     \textbf{(Training Datasets})  &        \textbf{\textit{S-140}}      &          \textbf{\textit{ACL}}      &         \textbf{\textit{IMDB}}     &      \textbf{\textit{YELP}}     \\ \hline
\multirow{2}{*}{} Softmax & \multirow{2}{*}{\textbf{83.08\%}} & \multirow{2}{*}{\textbf{82.90\%}} & \multirow{2}{*}{\textbf{79.90\%}} & \multirow{2}{*}{\textbf{84.20\%}} \\
                  (S-140) &                   &                   &                   &                   \\ \hline
\multirow{2}{*}{} Softmax & \multirow{2}{*}{81.70\%} & \multirow{2}{*}{69.50\%} & \multirow{2}{*}{64.50\%} & \multirow{2}{*}{79.60\%} \\
          (S-140 + Amazon)&                   &                   &                   &                   \\ \hline
\end{tabular}
\begin{tabular}{|c|c|c|c|c|}
\hline
\multirow{2}{*}{} \textbf{Our Models} & \multicolumn{4}{c|}{\textbf{Testing Datasets}}                                             \\ \cline{2-5} 
     \textbf{(Training Datasets})  &        \textbf{\textit{SST-1}}      &          \textbf{\textit{SST-2}}      &         \textbf{\textit{SWN-1}}     &      \textbf{\textit{SWN-2}}     \\ \hline
\multirow{2}{*}{} Tanh & \multirow{2}{*}{41.90\%} & \multirow{2}{*}{51.67\%} & \multirow{2}{*}{\textbf{15.77\%}} & \multirow{2}{*}{\textbf{66.64\%}} \\
                  (S-140) &                   &                   &                   &                   \\ \hline
\multirow{2}{*}{} Tanh & \multirow{2}{*}{\textbf{41.99\%}} & \multirow{2}{*}{\textbf{51.78\%}} & \multirow{2}{*}{14.30\%} & \multirow{2}{*}{60.40\%} \\
         (S-140 + Amazon) &                   &                   &                   &                   \\ \hline
\multirow{2}{*}{} \textbf{Our Models} & \multicolumn{4}{c|}{\textbf{Testing Datasets}}                                             \\ \cline{2-5}
     \textbf{(Training Datasets)}  &        \textbf{\textit{SST-1}}      &          \textbf{\textit{SST-2}}      &         \textbf{\textit{SWN-1}}     &      \textbf{\textit{SWN-2}}     \\ \hline
\multirow{2}{*}{} Softmax & \multirow{2}{*}{41.57\%} & \multirow{2}{*}{51.26\%} & \multirow{2}{*}{15.30\%} & \multirow{2}{*}{64.61\%} \\
                  (S-140) &                   &                   &                   &                   \\ \hline
\multirow{2}{*}{} Softmax & \multirow{2}{*}{39.92\%} & \multirow{2}{*}{49.22\%} & \multirow{2}{*}{14.20\%} & \multirow{2}{*}{59.97\%} \\
          (S-140 + Amazon)&                   &                   &                   &                   \\ \hline
\end{tabular}
\end{center}
\end{table}

\begin{table}
\begin{center}
\caption{Precision and Recall of our models against S-140 dataset.}\label{tab: tab3}
\begin{tabular}{|c|c|c|c|c|}
\hline
\multicolumn{5}{|c|}{S-140}                                                                            \\ \hline
\multirow{2}{*}{\textbf{}} Our Models & \multicolumn{2}{c|}{Positive}                 & \multicolumn{2}{c|}{Negative}                 \\ \cline{2-5} 
\textbf{(Training Datasets)}     &    Precision      &      Recall       &        Precision  &        Recall     \\ \hline
\multirow{2}{*}{} Tanh & \multirow{2}{*}{80.64\%} & \multirow{2}{*}{75.56\%} & \multirow{2}{*}{79.96\%} & \multirow{2}{*}{81.82\%} \\
               (S-140)   &                   &                   &                   &                   \\ \hline
\multirow{2}{*}{} Tanh & \multirow{2}{*}{82.34\%} & \multirow{2}{*}{\textbf{81.56\%}} & \multirow{2}{*}{81.38\%} & \multirow{2}{*}{82.55\%} \\
                (S-140 + Amazon)  &                   &                   &                   &                   \\ \hline
\multirow{2}{*}{} Softmax & \multirow{2}{*}{\textbf{83.73\%}} & \multirow{2}{*}{80.74\%} & \multirow{2}{*}{\textbf{81.57\%}} & \multirow{2}{*}{\textbf{85.42\%}} \\
                (S-140)  &                   &                   &                   &                   \\ \hline
\multirow{2}{*}{} Softmax & \multirow{2}{*}{83.40\%} & \multirow{2}{*}{79.20\%} & \multirow{2}{*}{80.16\%} & \multirow{2}{*}{84.20\%} \\
                (S-140 + Amazon)  &                   &                   &                   &                   \\ \hline
\end{tabular}
\end{center}
\end{table}

\begin{table}
\begin{center}
\caption{Precision and Recall of our models against ACL dataset.}\label{tab: tab4}
\begin{tabular}{|c|c|c|c|c|}
\hline
\multicolumn{5}{|c|}{ACL}                                                                            \\ \hline
\multirow{2}{*}{\textbf{}} Our Models & \multicolumn{2}{c|}{Positive}                 & \multicolumn{2}{c|}{Negative}                 \\ \cline{2-5} 
\textbf{(Training Datasets)}     &    Precision      &      Recall       &        Precision  &        Recall     \\ \hline
\multirow{2}{*}{} Tanh & \multirow{2}{*}{85.17\%} & \multirow{2}{*}{71.20\%} & \multirow{2}{*}{75.26\%} & \multirow{2}{*}{87.60\%} \\
               (S-140)   &                   &                   &                   &                   \\ \hline
\multirow{2}{*}{} Tanh & \multirow{2}{*}{80.65\%} & \multirow{2}{*}{75.00\%} & \multirow{2}{*}{76.64\%} & \multirow{2}{*}{82.00\%} \\
                (S-140 + Amazon)  &                   &                   &                   &                   \\ \hline
\multirow{2}{*}{} Softmax & \multirow{2}{*}{\textbf{87.64\%}} & \multirow{2}{*}{\textbf{76.60\%}} & \multirow{2}{*}{\textbf{79.22\%}} & \multirow{2}{*}{89.20\%} \\
                (S-140)  &                   &                   &                   &                   \\ \hline
\multirow{2}{*}{} Softmax & \multirow{2}{*}{85.45\%} & \multirow{2}{*}{47.00\%} & \multirow{2}{*}{63.45\%} & \multirow{2}{*}{\textbf{91.60\%}} \\
                (S-140 + Amazon)  &                   &                   &                   &                   \\ \hline
\end{tabular}
\end{center}
\end{table}

\begin{table}
\begin{center}
\caption{Precision and Recall of our models against IMDB dataset.}\label{tab: tab5}
\begin{tabular}{|c|c|c|c|c|}
\hline
\multicolumn{5}{|c|}{IMDB}                                                                            \\ \hline
\multirow{2}{*}{\textbf{}} Our Models & \multicolumn{2}{c|}{Positive}                 & \multicolumn{2}{c|}{Negative}                 \\ \cline{2-5} 
\textbf{(Training Datasets)}     &    Precision      &      Recall       &        Precision  &        Recall     \\ \hline
\multirow{2}{*}{} Tanh & \multirow{2}{*}{78.92\%} & \multirow{2}{*}{78.60\%} & \multirow{2}{*}{78.69\%} & \multirow{2}{*}{79.00\%} \\
               (S-140)   &                   &                   &                   &                   \\ \hline
\multirow{2}{*}{} Tanh & \multirow{2}{*}{71.53\%} & \multirow{2}{*}{77.40\%} & \multirow{2}{*}{75.38\%} & \multirow{2}{*}{69.20\%} \\
                (S-140 + Amazon)  &                   &                   &                   &                   \\ \hline
\multirow{2}{*}{} Softmax & \multirow{2}{*}{79.14\%} & \multirow{2}{*}{\textbf{81.20\%}} & \multirow{2}{*}{\textbf{80.70\%}} & \multirow{2}{*}{78.60\%} \\
                (S-140)  &                   &                   &                   &                   \\ \hline
\multirow{2}{*}{} Softmax & \multirow{2}{*}{\textbf{81.66\%}} & \multirow{2}{*}{37.40\%} & \multirow{2}{*}{59.40\%} & \multirow{2}{*}{\textbf{91.60\%}} \\
                (S-140 + Amazon)  &                   &                   &                   &                   \\ \hline
\end{tabular}
\end{center}
\end{table}

\begin{table}
\begin{center}
\caption{Precision and Recall of our models against YELP dataset.}\label{tab: tab6}
\begin{tabular}{|c|c|c|c|c|}
\hline
\multicolumn{5}{|c|}{YELP}                                                                            \\ \hline
\multirow{2}{*}{\textbf{}} Our Models & \multicolumn{2}{c|}{Positive}                 & \multicolumn{2}{c|}{Negative}                 \\ \cline{2-5} 
\textbf{(Training Datasets)}     &    Precision      &      Recall       &        Precision  &        Recall     \\ \hline
\multirow{2}{*}{} Tanh & \multirow{2}{*}{82.57\%} & \multirow{2}{*}{83.40\%} & \multirow{2}{*}{83.23\%} & \multirow{2}{*}{82.40\%} \\
               (S-140)   &                   &                   &                   &                   \\ \hline
\multirow{2}{*}{} Tanh & \multirow{2}{*}{82.40\%} & \multirow{2}{*}{82.40\%} & \multirow{2}{*}{82.40\%} & \multirow{2}{*}{82.40\%} \\
                (S-140 + Amazon)  &                   &                   &                   &                   \\ \hline
\multirow{2}{*}{} Softmax & \multirow{2}{*}{84.06\%} & \multirow{2}{*}{\textbf{84.40\%}} & \multirow{2}{*}{\textbf{84.34\%}} & \multirow{2}{*}{84.00\%} \\
                (S-140)  &                   &                   &                   &                   \\ \hline
\multirow{2}{*}{} Softmax & \multirow{2}{*}{\textbf{89.57\%}} & \multirow{2}{*}{67.00\%} & \multirow{2}{*}{73.64\%} & \multirow{2}{*}{\textbf{92.20\%}} \\
                (S-140 + Amazon)  &                   &                   &                   &                   \\ \hline
\end{tabular}
\end{center}
\end{table}

\begin{table}
\begin{center}
\caption{Precision and Recall of our models against SST dataset.}\label{tab: tab7}
\begin{tabular}{|c|c|c|c|c|}
\hline
\multicolumn{5}{|c|}{SST}                                                                            \\ \hline
\multirow{2}{*}{\textbf{}} Our Models & \multicolumn{2}{c|}{Positive}                 & \multicolumn{2}{c|}{Negative}                 \\ \cline{2-5} 
\textbf{(Training Datasets)}     &    Precision      &      Recall       &        Precision  &        Recall     \\ \hline
\multirow{2}{*}{} Tanh & \multirow{2}{*}{54.99\%} & \multirow{2}{*}{\textbf{64.45\%}} & \multirow{2}{*}{45.73\%} & \multirow{2}{*}{36.21\%} \\
               (S-140)   &                   &                   &                   &                   \\ \hline
\multirow{2}{*}{} Tanh & \multirow{2}{*}{\textbf{55.29\%}} & \multirow{2}{*}{62.17\%} & \multirow{2}{*}{\textbf{46.16\%}} & \multirow{2}{*}{39.21\%} \\
                (S-140 + Amazon)  &                   &                   &                   &                   \\ \hline
\multirow{2}{*}{} Softmax & \multirow{2}{*}{54.94\%} & \multirow{2}{*}{60.88\%} & \multirow{2}{*}{45.58\%} & \multirow{2}{*}{39.62\%} \\
                (S-140)  &                   &                   &                   &                   \\ \hline
\multirow{2}{*}{} Softmax & \multirow{2}{*}{54.22\%} & \multirow{2}{*}{46.39\%} & \multirow{2}{*}{44.82\%} & \multirow{2}{*}{\textbf{52.65\%}} \\
                (S-140 + Amazon)  &                   &                   &                   &                   \\ \hline
\end{tabular}
\end{center}
\end{table}

\begin{table}
\begin{center}
\caption{Precision and Recall of our models against SWN dataset.}\label{tab: tab8}
\begin{tabular}{|c|c|c|c|c|}
\hline
\multicolumn{5}{|c|}{SWN}                                                                            \\ \hline
\multirow{2}{*}{\textbf{}} Our Models & \multicolumn{2}{c|}{Positive}                 & \multicolumn{2}{c|}{Negative}                 \\ \cline{2-5} 
\textbf{(Training Datasets)}     &    Precision      &      Recall       &        Precision  &        Recall     \\ \hline
\multirow{2}{*}{} Tanh & \multirow{2}{*}{\textbf{65.50\%}} & \multirow{2}{*}{61.72\%} & \multirow{2}{*}{\textbf{67.55\%}} & \multirow{2}{*}{71.02\%} \\
               (S-140)   &                   &                   &                   &                   \\ \hline
\multirow{2}{*}{} Tanh & \multirow{2}{*}{58.26\%} & \multirow{2}{*}{56.34\%} & \multirow{2}{*}{62.19\%} & \multirow{2}{*}{64.02\%} \\
                (S-140 + Amazon)  &                   &                   &                   &                   \\ \hline
\multirow{2}{*}{} Softmax & \multirow{2}{*}{62.30\%} & \multirow{2}{*}{\textbf{63.12\%}} & \multirow{2}{*}{66.73\%} & \multirow{2}{*}{65.94\%} \\
                (S-140)  &                   &                   &                   &                   \\ \hline
\multirow{2}{*}{} Softmax & \multirow{2}{*}{63.08\%} & \multirow{2}{*}{36.31\%} & \multirow{2}{*}{58.81\%} & \multirow{2}{*}{\textbf{81.06\%}} \\
                (S-140 + Amazon)  &                   &                   &                   &                   \\ \hline
\end{tabular}
\end{center}
\end{table}

\begin{table}[]
\caption{Example Statement}\label{tab: tab9}
\begin{tabular}{|p{7cm}|p{1.1cm}|}
\hline
\multicolumn{1}{|c|}{Statement}  &   Score                \\ \hline
\multicolumn{2}{|c|}{Positivity Statement}                     \\ \hline
Wow... Loved this place.              &          1.0             \\ \hline
this place is good.                   &          0.99999             \\ \hline
Very very fun chef.                   &          0.99974             \\ \hline
\multicolumn{2}{|c|}{Negativity Statement}                     \\ \hline
Spend your money and time someplace else.              &          -0.99999             \\ \hline
Overpriced for what you are getting.                   &          -0.99833             \\ \hline
Hell no will I go back.                   &          -1.0            \\ \hline
\multicolumn{2}{|c|}{Objectivity Statement}                     \\ \hline
It was a bit too sweet, not really spicy enough.   & 0.03396    \\ \hline
\multirow{2}{*}{}  They could serve it with just the vinaigrette and it may   & \multirow{2}{*}{0.00032}     \\
                    make for a better overall dish.     &                       \\ \hline
\multirow{2}{*}{}  They also now serve Indian naan bread with hummus and     & \multirow{2}{*}{0.00192}    \\
                   some spicy pine nut sauce that was out of this world.    &                       \\ \hline
\end{tabular}
\end{table}

\begin{table}
\begin{center}
\caption{Precision of our models against neutral statement dataset.}\label{tab: tab10}
\begin{tabular}{|c|c|c|c|c|c|}
\multicolumn{5}{c}{}                                                                            \\ \hline
\multirow{2}{*}{\textbf{}} Amazon product & \multicolumn{2}{c|}{tanh} & \multicolumn{2}{c|}{softmax} \\ \cline{2-5} 
\textbf{(reviews \& metadata)}     &    Neutral &     All &   Neutral &     All \\ \hline
\multirow{2}{*}{Cell Phones } & \multirow{2}{*}{\textbf{42.80\%}} & \multirow{2}{*}{82.60\%} & \multirow{2}{*}{\textbf{16.40\%}} & \multirow{2}{*}{77.10\%} \\
                  &                   &                   &                   &                   \\ \hline
\multirow{2}{*}{Musical Instruments}  & \multirow{2}{*}{39.50\%} & \multirow{2}{*}{\textbf{84.40\%}} & \multirow{2}{*}{10.9\%} & \multirow{2}{*}{\textbf{87.10\%}} \\
                  &                   &                   &                   &                   \\ \hline
\multirow{2}{*}{Home and Kitchen} & \multirow{2}{*}{35.30\%} & \multirow{2}{*}{82.90\%} & \multirow{2}{*}{15\%} & \multirow{2}{*}{81.90\%} \\
                  &                   &                   &                   &                   \\ \hline
\multirow{2}{*}{Toys and Games} & \multirow{2}{*}{29.90\%} & \multirow{2}{*}{84.20\%} & \multirow{2}{*}{12.80\%} & \multirow{2}{*}{83.80\%} \\
                  &                   &                   &                   &                   \\ \hline
\multirow{2}{*}{Office Products}  & \multirow{2}{*}{31.40\%} & \multirow{2}{*}{81.90\%} & \multirow{2}{*}{10.70\%} & \multirow{2}{*}{84.20\%} \\
                  &                   &                   &                   &                   \\ \hline
\multirow{2}{*}{Sports and Outdoors}  & \multirow{2}{*}{37.70\%} & \multirow{2}{*}{82.40\%} & \multirow{2}{*}{12.60\%} & \multirow{2}{*}{84.40\%} \\
                  &                   &                   &                   &                   \\ \hline
\end{tabular}
\end{center}
\end{table}

\begin{table}
\begin{center}
\caption{The datasets of our web crawler.}\label{tab: tab11}
\begin{tabular}{|c|c|c|c|c|c|}
\hline
Date & Twitter & FB Post & FB Comment & Telegram & Total\\ \hline
11/04 & 4169 & 285 & 699 & 173704 & 178857 \\ \hline
11/05 & 6208 & 653 & 1076 & 166495 & 174432 \\ \hline
11/06 & 7317 & 657 & 1140 & 227293 & 236407 \\ \hline
11/07 & 3371 & 724 & 1258 & 231893 & 237246 \\ \hline
11/08 & 4577 & 708 & 1233 & 230196 & 236714 \\ \hline
11/09 & 6868 & 631 & 1170 & 234593 & 243262 \\ \hline
11/10 & 5361 & 362 & 950 & 221218 & 227891 \\ \hline
Average & 5438 & 572 & 1065 & 208669 & 215745 \\ \hline
Total & 2794325 & 205354 & 896913 & 53566665 & 57463257 \\ \hline
\end{tabular}
\end{center}
\end{table}

\section{Related Work}
The sentiment of a sentence can be inferred with subjectivity classification and polarity classification, where the former classifies whether a sentence is subjective or objective and the latter decides whether a subjective sentence expresses a negative or positive sentiment. In existing deep learning models, sentence sentiment classification is usually formulated as a joint three-way classification problem, namely, to predict a sentence as positive, neutral, and negative. Wang et al. proposed a regional CNN-LSTM model, which consists of two parts: regional CNN and LSTM, to predict the valence arousal ratings of text \cite{wang}. Wang et al. described a joint CNN and RNN architecture for sentiment classification of short texts, which takes advantage of the coarse-grained local features generated by CNN and long-distance dependencies learned via RNN \cite{wang2}. Guggilla et al. presented a LSTM and CNN-based deep neural network model, which utilizes word2vec and linguistic embeddings for claim classification (classifying sentences to be factual or feeling) \cite{Guggilla}. Kim also proposed to use CNN for sentence-level sentiment classification and experimented with several variants, namely CNN-rand (where word embeddings are randomly initialized), CNN-static (where word embeddings are pre-trained and fixed), CNN-non-static (where word embeddings are pre-trained and fine-tuned) and CNN-multichannel (where multiple sets of word embeddings are used)\cite{kim}.

Besides, about using machine learning to solve fraud problems, Fabrizio Carcillo et. al proposed to reduce credit card fraud by using machine learning techniques to enhance traditional active learning strategies \cite{DSAA}. Shuhao Wang et. al observes users’ time series data in the entire browsing sequence and then uses in-depth learning methods to model the detection of e-commerce fraud \cite{Cham}. As blockchain and virtual currencies prevail, fraudulent transactions cannot be ignored. Toyoda et al. identify characteristics of fraudulent Bitcoin addresses by extracting features from transactions through analysis of transaction patterns \cite{GLOBECOM}; Bian et. al predict the quality of ICO projects through different kinds of information such as white papers, founding teams, GitHub libraries, and websites \cite{ICORating}.

\section{Conclusion}
We developed a LSTM + CNN system for detecting the sentiment analysis of social-network opinion in this paper. Compared with the baseline approach, our system obtains good results. Our goal is to optimize the amount of parameters, network structure, and release automated detection tools, public RESTful API and chatbot. The future work is to reduce the complex task and train for higher performance in confronting the sentiment, in order to improve sentiment analysis. We also built a "help us" website to label more datasets is shown in Fig. \ref{fig: help_us}\footnote{http://soc.twman.org}. The experiment material and research results are shown on the website if there are any updates.

More importantly, we collected the users’ cryptocurrency comments from the social network (as shown in Table \ref{tab: tab11}) and deploy our sentiment analysis on RatingToken and Coin Master (Android application of Cheetah Mobile Blockchain Security Center). Our proposed methodology can effectively provide detail information to resolve risks of being fake and fraud problems. Fig. \ref{fig: SOC002} is the screenshot of our sentiment analysis production website.

\section*{Acknowledgement}
This work would not have been possible without the valuable dataset offered by Cheetah Mobile. Special thanks to RatingToken and Coin Master.

\begin{figure}
	\centering
	\includegraphics[width=3.2in,height=1.5in]{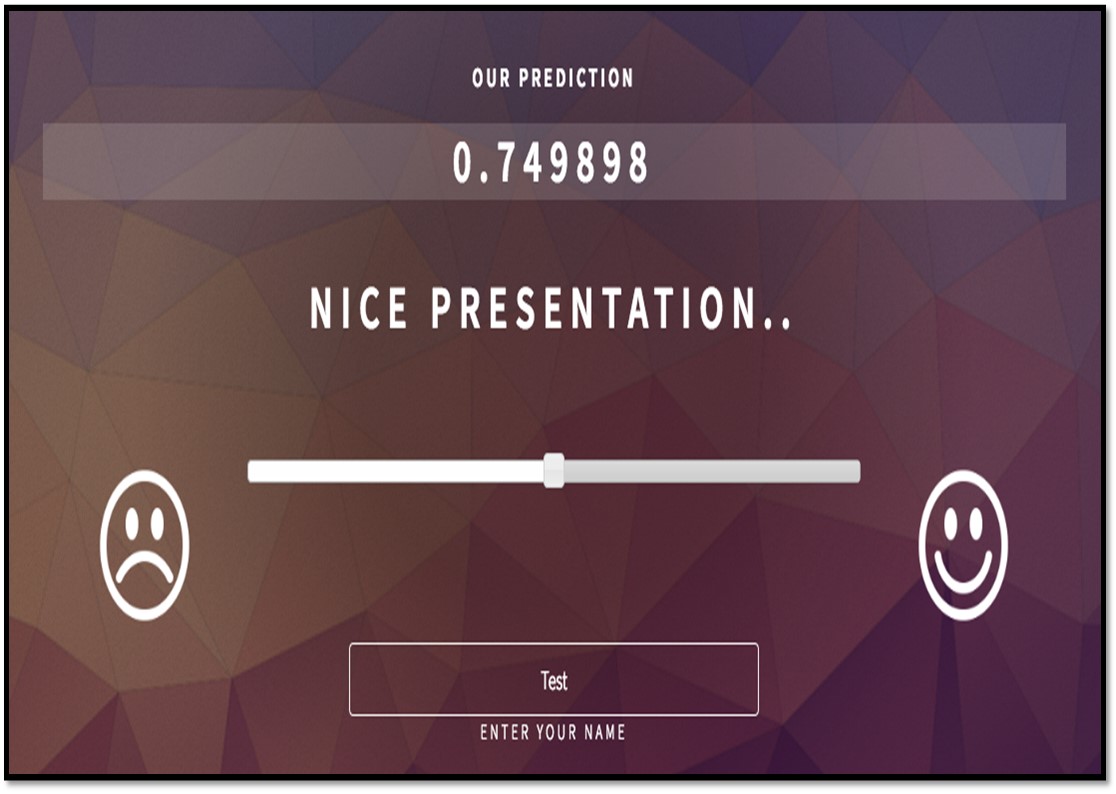}
	\caption{The screenshot of our help us website.}\label{fig: help_us}
\end{figure}

\begin{figure}
	\centering
	\includegraphics[width=3.2in,height=4.2in]{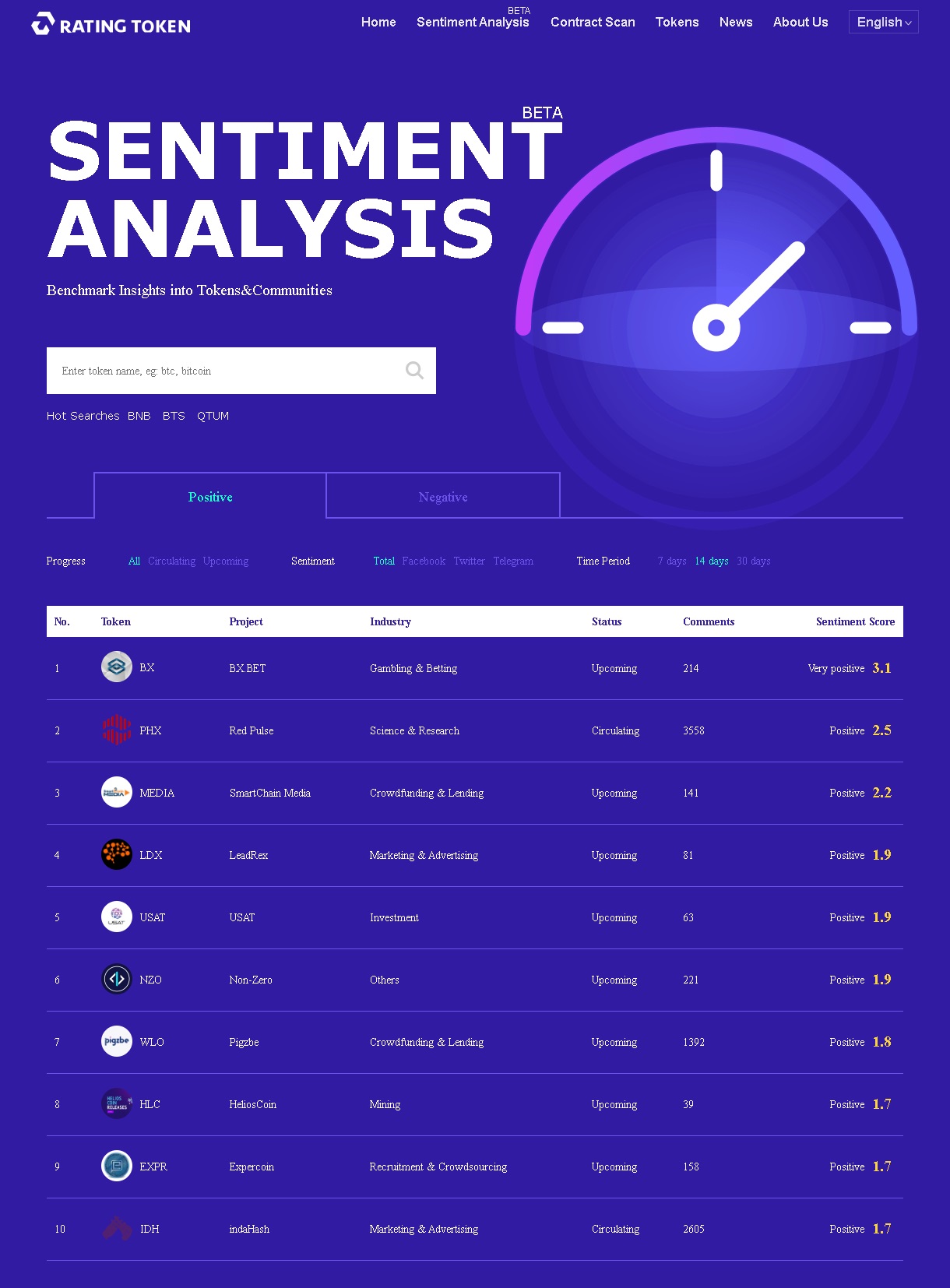}
	\caption{The screenshot of our sentiment analysis production website.}\label{fig: SOC002}
\end{figure}


\bibliographystyle{ACM-Reference-Format}

\begin{thebibliography}{99}
\bibitem{bitcoin}
S. Nakamoto. Bitcoin: A peer-to-peer electronic cash system. http://bitcoin.org/bitcoin.pdf.

\bibitem{surveyvulnerability}
X. Li, P. Jiang, T. Chen, X. Luo, and Q. Wen, "A survey on the security of blockchain systems," Future Generation Computer Systems,arxiv: 1802.06993

\bibitem{BitcoinMagazine}
Vitalik Buterin, "A Next-Generation Cryptocurrency and Decentralized Application Platform," Bitcoin Magazine, [Online]. Available: https://bitcoinmagazine.com/articles/ethereum-next-generation-cryptocurrency-decentralized-application-platform-1390528211/

\bibitem{Pennsylvania}
Cohney, Shaanan and Hoffman, David A. and Sklaroff, Jeremy and Wishnick, David A., "Coin-Operated Capitalism". Columbia Law Review, Forthcoming.

\bibitem{glove}
J. Pennington, R. Socher, and C. D. Manning. Glove: Global vectors for word representation. In Empirical Methods in Natural Language Processing (EMNLP), pages 1532–1543, 2014.

\bibitem{LSTM}
S. Hochreiter and J. Schmidhuber. Long short-term memory. Neural computation, 9(8):1735–1780, 1997.

\bibitem{SELU}
G. Klambauer, T. Unterthiner, A. Mayr, and S. Hochreiter. Self-normalizing neural networks. 31st Conference on Neural Information Processing Systems, NIPS 2017, 2017.

\bibitem{Adam}
D. Kingma and J. Ba, "Adam: A method for stochastic optimization," in International Conference for Learning Representations, 2015.

\bibitem{sentiment140}
Go, A., Bhayani, R. and Huang, L., 2009. Twitter sentiment classification using distant supervision. CS224N Project Report, Stanford, 1(2009), p.12.

\bibitem{kim}
Y. Kim, "Convolutional Neural Networks for Sentence Classification", Proceedings of the 2014 Conference on Empirical Methods in Natural Language Processing (EMNLP 2014), pp. 1746-1751, 2014.

\bibitem{SWN}
Esuli, A., Sebastiani, F. 2006. SentiWordNet: A publicly available lexical resource for opinion mining. In: Proc. of LREC.

\bibitem{wang}
Wang, Jin, et al. "Dimensional sentiment analysis using a regional cnn-lstm model." The 54th Annual Meeting of the Association for Computational Linguistics. Vol. 225, 2016.

\bibitem{wang2}
Wang X, Liu Y, Sun C, Wang B, and Wang X. Predicting polarities of tweets by composing word embeddings with long short-term memory. In Proceedings of the Annual Meeting of the Association for Computational Linguistics (ACL 2015), 2015.

\bibitem{Guggilla}
Guggilla C, Miller T, Gurevych I. CNN-and LSTM-based claim classification in online user comments. In Proceedings of the International Conference on Computational Linguistics (COLING 2016), 2016.

\bibitem{DSAA}
F. Carcillo, Y. A. L. Borgne, O. Caelen, and G. Bontempi, "An Assessment of Streaming Active Learning Strategies for \newpage Real-Life Credit Card Fraud Detection," in 2017 IEEE International Conference on Data Science and Advanced Analytics (DSAA), 2017, pp. 631-639.

\bibitem{Cham}
S. Wang, C. Liu, X. Gao, H. Qu, and W. Xu, "Session-Based Fraud Detection in Online E-Commerce Transactions Using Recurrent Neural Networks," in Machine Learning and Knowledge Discovery in Databases, Cham, 2017, pp. 241-252.

\bibitem{GLOBECOM}
K. Toyoda, T. Ohtsuki, and P. T. Mathiopoulos, "Identification of High Yielding Investment Programs in Bitcoin via Transactions Pattern Analysis," in GLOBECOM 2017 - 2017 IEEE Global Communications Conference, 2017, pp. 1-6.

\bibitem{ICORating}
S. Bian, Z. Deng, F. Li, W. Monroe, P. Shi, Z. Sun, et al., "IcoRating: A Deep-Learning System for Scam ICO Identification," arXiv:1803.03670, 2018.

\end{thebibliography}

\end{document}